\documentclass[fleqn,10pt,twocolumn]{AROB-ISBC-SWARM22}

\usepackage{algpseudocode}
\usepackage{amsmath}
\usepackage{algorithm}

\title{Energy-Aware Multi-Robot Task Allocation in Persistent Tasks}

\author{Ehsan Latif, Yikang Gui, Aiman Munir, Ramviyas Parasuraman}
\speaker{Ehsan Latif}

\affils{Department of Computer Science, University of Georgia, Athens, GA 30602, United States\\
(Corresponding author e-mail: {ramviyas}@uga.edu)
}
\abstract{%
The applicability of the swarm robots to perform foraging tasks is inspired by their compact size and cost. A considerable amount of energy is required to perform such tasks, especially if the tasks are continuous and/or repetitive. Real-world situations in which robots perform tasks continuously while staying alive (survivability) and maximizing production (performance) require energy awareness. This paper proposes an energy-conscious distributed task allocation algorithm to solve continuous tasks (e.g., unlimited foraging) for cooperative robots to achieve highly effective missions.
We consider efficiency as a function of the energy consumed by the robot during exploration and collection when food is returned to the collection bin. Finally, the proposed energy-efficient algorithm minimizes the total transit time to the charging station and time consumed while recharging and maximizes the robot's lifetime to perform maximum tasks to enhance the overall efficiency of collaborative robots. We evaluated the proposed solution against a typical greedy benchmarking strategy (assigning the closest collection bin to the available robot and recharging the robot at maximum) for efficiency and performance in various scenarios. The proposed approach significantly improved performance and efficiency over the baseline approach.}


\begin{document}

\maketitle


\section{Introduction}

The ability to coordinate and self-organize in swarm robots are the main features that draw inspiration from natural social systems such as ants colonies \cite{lenka2018}. They allow a large group of artificial agents (robots) to achieve collective intelligence and efficiency that individual robots cannot achieve. In a swarm of robots, each robot builds knowledge of the local world through its limited perception. However, robots exploit direct and indirect communication to increase their efficiency, such as groups by sharing information and making decisions together \cite{yang2019self}. 

One well-researched example is the application of swarms for foraging \cite{winfield2009towards}, which is a behavior commonly observed in social animals. This task is widely studied among robot swarms due to its importance as a metaphor for a wide range of robotic applications, including search and rescue and resource extraction (e.g., harvesting) \cite{liu2007,campo2007efficient,yang2020needs}.

\begin{figure*}[t]
\centering
 \includegraphics[width=\linewidth]{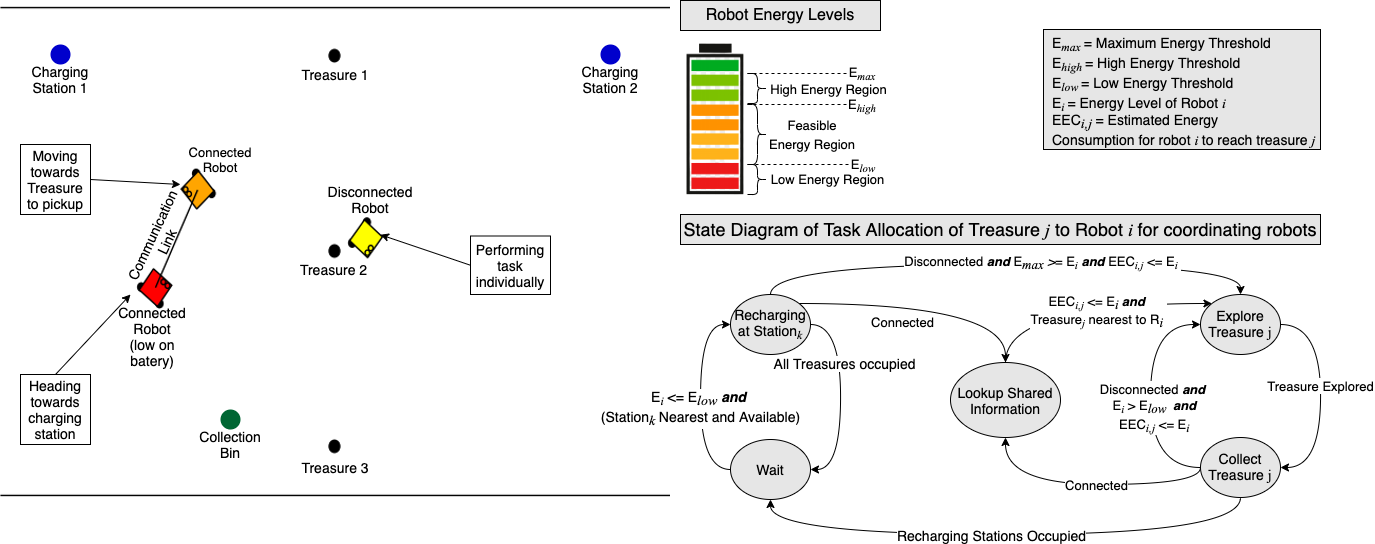}
 \caption{Overview of the proposed energy-aware task allocation algorithm, with the robot state diagram.}
 \label{fig:overview}
 \hspace{-8mm}
\end{figure*}

In foraging, the robots search extensively so that resources are brought back to a central ("nest") location \cite{erusagounder2018}. When comparing foraging methods to determine which is best for a particular application, some desired functions are simplicity, scalability, decentralization, discovery, and parallelism. 
Systems with these attributes are much better equipped to handle applications with multipurpose in dynamic environments, such as disaster relief and survey \cite{nauta2020}.
 
In most previous works, the measure of performance was the time spent foraging, which does not cover the full cost required to complete the quest, as it ignores the actual energy used or stored and the time it takes to recharge a given robot. While teams of robots need to maintain optimized speed and efficiency, time is not the only measure of energy consumption \cite{notomista2021}. Another measure is the remaining battery energy level in each robot \cite{chen2019}. It is essential to maintain energy in the robot and efficiently deplete energy while feeding. Assuming an infinite source of energy for the robot, which can be billed at  100$\%$ battery level regardless of the actual amount of energy required, is not realistically possible \cite{mayya2019,parasuraman2012energy}.
Moreover, the robot team can also be heterogeneous, meaning the battery capacities and speed can vary within the system \cite{ov2020impact,yang2020hierarchical}.

It makes no sense to load a robot to its maximum capacity if it knows to have resources near the nest. This robot can use the rest of the energy needed to complete these simple tasks. Other robots which struggle to find resources in other areas will have to be careful about being recharged and not use energy more than their maximum capacity \cite{zedadra2015}.
Very few papers consider battery power when measuring energy efficiency, so when they try to translate algorithms to the real world, their algorithms suffer from this limitation. 
 
Previous work (e.g., \cite{campo2007efficient}) has focused only on measuring energy during feeding and has not taken into account the unused energy in each robot's battery. Thus, the overall performance would be less efficient than an energy-aware swarm system.  The most commonly used metrics for system cost are the search and retrieval times.
Therefore, an additional indicator should include a measure of the remaining energy in each robot battery. No matter how well the foraging task performs, systems must be sensitive to energy because the 'income' of foraging (i.e., how many items are collected in a certain quantity of time) should be greater than the cost to get it. 
The system cost can be calculated with respect to the energy consumption conditions of the swarm.
 
This paper proposes an energy-aware distributed task allocation algorithm to solve persistently occurring tasks (e.g., unlimited foraging) applied to cooperative robots with limited energy capacity. The algorithm aims to achieve high efficiency in foraging while keeping the robots alive without depletion of total energy and networking (data communication). We consider efficiency as a function of the energy consumed by the robots during exploration and collection when the food is returned to the collection bin. Our algorithm introduces energy-level thresholds for different tasks, including autonomous recharging, a communication mechanism between the robots to stay aware and networked, and a deadlock avoidance algorithm to avoid them colliding while keeping them from being stuck in a deadlock state in terms of task allocation.

We consider practical limitations on the resources, such as limited recharging stations and fewer robots available than the number of tasks. The novelty of the proposed algorithm can be summarized as follows:
\begin{enumerate}
    \item The proposed algorithm strives to find a balance between recharging and foraging tasks in a distributed manner.
    \item The algorithm also ensures that each robot only gets charged when necessary while allowing others to do the same. Ultimately, the energy-aware algorithm minimizes the sum of time to transit to the charging station and the time consumed during charging while maximizing the survival time of a robot to perform a maximum number of tasks to improve the overall efficiency. 
    \item Additionally, we devise a deadlock avoidance strategy integrated into our algorithm by exploiting the randomization gain in the task allocation decision-making process to tackle the deadlock scenario, where robots may indefinitely wait to avoid collisions. 
    \item Finally, the algorithm maintains connectivity between the robots to enable them to communicate effectively throughout the task deployment period, such that the overall system performance is enhanced.
\end{enumerate}
 
An overview of the proposed solution is depicted in Fig.~\ref{fig:overview}. The solution has been extensively analyzed with various agents and food sources using the Robotarium platform, a remotely accessible simulator-hardware integrated test-bed for swarm robotics research \cite{pickem2016}. We have performed experiments to validate the algorithm in terms of its persistence, that is, how many tasks the robot can efficiently complete while remaining alive. We evaluated the proposed solution compared with a typical greedy-based reference strategy (which assigns the closest treasure to the available robot and recharges the robot to the maximum level when depleted) in terms of efficiency and performance in different scenarios. The efficiency function depends on the following parameters: the number of alive robots, total traveled distance, time spent on recharging, number and value of collected treasures, and connectivity maintenance. 

The proposed approach has demonstrated a significant improvement in performance and efficiency than the greedy reference approach. The simulation results confirm the robustness and practicality of the proposed solution in persistent tasks.
The rest of the paper is organized as follows. Sec.~\ref{sec:approach} introduces the problem statement and elaborates the algorithm, Sec.~\ref{sec:results} describes the simulation setup, experimentation, and the results. In Sec.~\ref{sec:conclusion}, we conclude the paper.

\section{Proposed Approach}
\label{sec:approach}
\textbf{Problem Statement} Given a set of robots $R$, a set of treasures $T$, a set of significance values $V$ associated to specific treasure, and a set of cost of treasure $C$, the goal is to maximize the value (v) and minimize the cost (c) made by assigning a robot to treasure $t$. Set of Robots $R = \{r_1, r_2, . . , r_n \}$ where n = number of robots, set of Treasures $T = \{ t_1, t_2, . . , t_k \}$, where k = number of treasures, significance values of treasures are predefined and fixed, i.e; $V = \{ v_1, v_2, . . , v_k \}$, where $v_k$ is value of $t_k$, Cost associated to each treasure is $C = \{c_1,c_2, .. , c_k \}$, where $c_k$ = battery consumption to pick treasure $k$ from its location and drop to the collection bin. Set of robots’ current battery status $B = \{b_1, b_2, . . , b_n \}$, where $b_n$ = $n^{th}$ robot's battery level. Set of energy requirement to reach treasure $\tau = \{ \tau_{11}, \tau_{12}, .. , \tau_{1k}, . . , \tau_{nk} \}$, where $\tau_{nk}$ = energy required to pick treasure $k$ by robot $n$.
The battery consumption as a cost $c_i$ of treasure ($t_i$) can be calculated as:
\begin{equation}
    \centering
    c_i = (\alpha \times 1 + \beta \times distance(t) + \gamma) .
\end{equation}
Treasure utility function will be calculated as:
\begin{equation}
    \centering
    U_{i j} = v_j - ( \tau_{i j} + c_j ),
    \label{eq:util}
\end{equation}
where $1\leq i \leq n$ and $1\leq j \leq k$ and $b_i \geq (\tau_{i j} + c_j )$.

To solve this problem, we propose an auction-based energy-aware approach that takes care of each robot's four critical aspects in a distributed manner: 1) recharge planning; 2) treasure task allocation; 3) deadlock avoidance; and 4) connectivity maintenance.

\subsection{ Recharging Planner}
The intuition behind the energy-Aware solution is that the robots should not go to recharge until the last moment (e.e., before reaching a dangerous state where the battery may die and the robot will become useless). 
The algorithm solves how to predict the 'last moment' and how to arrange the robots to recharge. Accordingly, we will set the following parameters to help us:
\begin{itemize}
\item no-recharger: the number of recharge stations.
\item no-robot: the number of robots.
\item id: the id of a robot
\item current-energy-level: the battery level for each robot.
\item avg-task-energy: the average energy needed for one task.
\item avg-task-time: the average time for one task.
\item avg-recharge-time: the average time for recharging.
\item energy-threshold: the battery threshold for a robot before it must be recharged.
\end{itemize}

Algorithm \ref{alg:recharge} plan the recharging based on the following objective to minimize the time travel to capture treasure and return to the collection bin while keeping the robot alive subject to the battery constraint.

\begin{algorithm}
\caption{Energy-Aware Recharging Planner}
\label{alg:recharge}
\begin{algorithmic}[1]
\State predict-episode = ceil(no-robot / no-recharger)
\State task-recharge-time-ratio = ceil(avg-recharge-time / avg-task-time)
\State predict-energy = list()
\State recharging-plan-robot-ids = set()
\State i=0
\While {$i \leq predict-episode$}
    \State predict-energy.append(current-energy-level - (i * task-recharge-time-ratio * avg-task-energy))
    \State i++
    \For {index in length(predict-energy) }
        \If {$index * no-recharger < predict-energy[i] \leq index * no-recharger$}
            \If {recahrging-station-available}
                \State add the id of the robot with the lowest energy level to recharging-plan-robot-ids
            \EndIf
        \EndIf
    \EndFor
\EndWhile
\State return recharging-plan-robot-ids
\end{algorithmic}
\end{algorithm}

While analyzing the energy-aware recharging planner, steps 6-9 in the algorithm will guarantee that the robot will never die, resulting in the maximum usage rate of the recharge stations. On the other hand, steps 9-12 in the algorithm will guarantee that the robots will never go to the recharge station until the last moment. The number of recharging will be minimum, leading to the minimum number of times the robot will travel to recharging stations.

\subsection{Task Allocation}
A robot can assume one of the four states at a time: 1) idle; 2) recharging 3) pickup treasure 4) going to recharge/treasure location. When a robot $r$ gets to idle state, it computes its utility function $U_{rt}$ against each available treasure $t$ from the set $T$. It then finds its suitability for a treasure $t$ by finding the maximum $U_{rt}$ for a treasure $t$. As we can only communicate to the neighboring robots from the set $R$ in the connected graph, each idle robot $r$ broadcasts its maximum utility $U_{rt}$ against a treasure $t$ to all neighboring robots. Through which each idle robot at a specific time gets the maximum utility $U_{rt}$ of robot $r$ for a treasure $t$  and achieves consensus in the task assignments until they reach no changes in their utility values.

Here, each robot $r$ then picks the treasure $t$ as: if $U_{rt} > \forall U_{it}$, then robot $r$ assigns treasure $t$ else it repeats the previous step and excludes the treasure $t$ until it is assigned a treasure where $1 \leq i \leq$ number of idle robots and $i \neq r$.

In the case of a disconnected robot (from the communication graph/network), $d_r$ computes its utility $U_{drt}$ and assigns itself the treasure with the highest utility. There could be a possibility in which two robots are assigned to the same treasure if they were disconnected. Therefore, whenever this robot gets reconnected to the other robots (through the connectivity maintenance algorithm in Sec.~\ref{sec:connectivity}), it recomputes its suitability to the treasure based on neighboring robots' utilities. This whole process is explained in Algorithm~\ref{alg:task}, which assigns the task to an individual robot based on its energy level and availability of treasure.

\begin{algorithm}
\caption{Energy-Aware Task Allocation}\label{alg:task}
\begin{algorithmic}[1]
\While{True}
    \State robot-recharging-ids =  recharging-planner (Alg. ~\ref{alg:recharge})
    \For{$r$ in $R$}
        \If{r in  robot-recharging-ids}
            \State $task_r$ = recharging
        \ElsIf{r is connected and idle}
            \For{t in treasures}
                \State Compute $U_{rt}$ as (Eq.~\ref{eq:util})
            \EndFor
            \State assign a treasure with maximum $U_{rt}$
        \EndIf
    \EndFor
\EndWhile
\end{algorithmic}
\end{algorithm}

\subsection{Deadlock Avoidance}
The proposed approach also ensures the continuous movement of robots while avoiding collision by incorporating the deadlock avoidance technique. Due to the usage of Robotarium's inbuilt collision avoidance algorithm that changes the robot's trajectory to avoid collisions, the robots can be in a deadlock scenario without movement possibility. Here we provide a simple technique to solve the problem.

First, we need to identify whether a robot is in a deadlock scenario. The deadlock detection algorithm works so that if the distance between two robots is less than some distance threshold for a specified time, we will mark these two robots deadlocked. 
Second, once two robots are detected deadlocked, we will use the following deadlock avoidance algorithm to free them.

Suppose robots A and B are in a deadlock scenario. Robot A should move in the opposite direction of Robot B plus some random noise (alpha angle). Robot B should move in the opposite direction of Robot A plus some random noise (alpha angle). The random noise here is to avoid the exact opposite direction because, in the experiments, the exact opposite direction is not helpful to avoid the deadlock scenario. The reason behind it is the mechanism of the unicycle robot in the Robotarium platform \cite{pickem2017robotarium}.

\subsection{Connectivity Maintenance}
\label{sec:connectivity}

The objective of the integration of connectivity maintenance in this energy-aware task allocation algorithm is three-fold: 1) increase the graph connectivity so that more robots can communicate and share data at any given time; 2) keep the robots alive in terms of connectivity, and 3) maximize the number of foraging tasks completed by keeping a connected graph all the time.

The maintenance approach works in the following way. When the robot gets disconnected, after completing the currently assigned task, it moves closer to the robot at a minimum distance using its knowledge of the closest robot's state before disconnection. The assumption is that once this robot reaches this new position, there is a guarantee of re-connection since the previously-closest robot would not have moved beyond its communication radius within this short interval of time. As soon as the robot gets reconnected, it can execute the next task through the auction mechanism in Alg.~\ref{alg:task}.
The maximum priority is to keep the robots alive, both energetically and networking-wise, so they persistently execute the task, resulting in improved system efficiency and performance.

\begin{table*}[]
\resizebox{\textwidth}{!}{%
\begin{tabular}{|l|l|l|l|}
\hline
\textbf{Performance Indices}                & \textbf{Baseline} & \textbf{Proposed w/o deadlock avoidance} & \textbf{Proposed w/ deadlock avoidance} \\ \hline \hline
Average number of alive robots (units)      & 3.6               & 4                                       & 4.8                                     \\ \hline
Total distance traveled (m)                 & 160.2             & 153.27                                   & 170.39                                  \\ \hline
Go-To recharging time (sec)                 & 3057.6            & 3523.4                                   & 4539.3                                  \\ \hline
Recharging time (sec)                       & 4060.7            & 6213.7                                   & 5226.16                                 \\ \hline
Wait-For recharging time (sec)              & 67.98             & 66.2                                     & 52.12                                   \\ \hline
Total number of treasures collected (units) & 51.6              & 61.9                                     & 65.9                                    \\ \hline
Total treasure value achieved (units)       & 251.6             & 292.1                                    & 357.9                                   \\ \hline
\end{tabular}}
\caption{Performance comparison of the proposed approach with the baseline and without deadlock avoidance strategies.}
\label{table:results}
\end{table*}
\begin{table*}[]
\resizebox{\textwidth}{!}{%
\begin{tabular}{|l|l|l|l|}
\hline
\textbf{Performance Indices}                & \textbf{Proposed w/o connectivity maintenance} & \textbf{Proposed w/  connectivity maintenance} \\ \hline \hline
Average number of alive robots (units)       & 4                & 4.8 \\ \hline
Total distance traveled (m)                  & 213               & 166 \\ \hline
Go-To recharging time (sec)                  & 3214.2            & 4512.4  \\ \hline
Recharging time (sec)                       & 4162.2            & 5213.5  \\ \hline
Wait-For recharging time (sec)              & 57.28             & 64.7   \\ \hline
Total number of treasures collected (units) & 61.4              & 63.5  \\ \hline
Total treasure value achieved (units)       & 251.6             & 292.1  \\ \hline
\end{tabular}}
\caption{Performance comparison of the proposed approach with and without connectivity maintenance.}
\label{table:result1}
\end{table*}

\section{Experimentation and Results}
\label{sec:results}
The efficiency and effectiveness of the proposed approach have been validated through extensive simulation experiments on the Robotarium platform, which is an open-source tool enabling quick validation of multi-robot algorithms. We have set up the Robotarium environment for five collaborating robots, five treasures at fixed positions, five collection bins, and two discrete recharging stations on the same testbed can be seen in Fig\ref{fig:overview}. We have implemented the below greedy-based baseline task allocation approach in comparison to our proposed approach.

\textbf{Baseline} We have used greedy task allocation as a baseline strategy which works in such a way that for all available robots, it assigns the task at the nearest location. In case of more than one robot is assigned for a task location, it assigns the robot with the shortest distance for this task. The same procedure is repeated until all available robots are assigned a unique task location. Moreover, if at least one robot is available, it can then repeat this procedure. This strategy sets the low battery threshold to 30\% for every robot. When any robot with a battery level is below the threshold, it assigns that robot the available recharging station. If both recharging stations are taken, then it waits until one is available (while it waits that the robot is not available for task allocation). It also sets the maximum charging threshold to 60\%, and When the robot energy reaches the threshold (after putting in a recharging station), that robot becomes available for task allocation. Once the task is assigned, the robot moves towards the task location and then approaches the collection point; once it is done with treasure collection, it makes this robot available for the next task.

\textbf{Parameters} To perform the experiments, we have set the following simulation parameters.
\begin{itemize}
    \item Number of robots: the number of robots that participate in each simulation.
    \item Treasures: the number of treasures (in units) that a robot can transport at each time (robot capacity) and its value.
    \item Simulation Space: include the dimension of the search space, which is a grid of $N \times N$ cells (world
size). 
    \item World complexity, which can be obstacle-free or obstacle environment. 
    \item Sink number, which is the number of collection stations to where agents return food.
    \item Recharging stations, where robots get themselves recharge.
    \item Treasures: the number of treasure locations (food density) that are located at fixed positions.
    \item $E_{max}$ is the maximum value that a robot can recharge at the recharging station.
    \item $E_c$ represents the current energy that a robot will have at the time (t). It is set initially to $E_{max}$ that is 100\%. 
    \item $E_{min}$ is the threshold that activates the return to home and recharges energy behavior which is fixed, and in our case, it is 20\%. When $E_c \leq E_{min}$, the robot needs to move towards the recharging station to recharge energy.
\end{itemize}

\textbf{Performance Indices} To evaluate the performance of the algorithm, we have used the following indices:
\begin{itemize}
    \item Average number of alive robots,
    \item Total distance traveled.
    \item Go-To recharging time,
    \item Wait-For recharging time,
    \item Total Recharging time,
    \item Total number of treasures collected,
    \item Total treasure value achieved
\end{itemize}

\textbf{Energy Models}
A robot will lose energy based on the following energy consumption model \cite{parasuraman2014model}:
\begin{equation}
    \centering
    E_t = E_{t-1} - (\alpha \times 1 + \beta \times distance(t) + \gamma \times \text{isTreasurePicked}),
\end{equation}
where $\alpha$ represents the coefficient for the static energy consumption due to onboard computing and sensors of the robot for example, $\beta$ represents the coefficient for the dynamic power consumption by the robot due to movement (e.g, velocity-based motor power to move to a specific distance), and $\gamma$ represents the coefficient for the energy spent in picking up the treasure (e.g., manipulation effort).

Similarly, a robot will gain energy if it is at the recharging station with the following model:
\begin{equation}
    \centering
E_t = E_{t-1} +\delta\times 1,
\end{equation}
where $\delta$ represents the rate at which the battery can be recharged. In our experiments, we use the following values: $\alpha = 0.1$, $\beta=2$, $\gamma=0.1$, $\delta =0.5$.
Initially, at time $t=0$, we start all the robots with full energy $E_0 = 100\%$.

\subsection{Results and Discussion}
We have performed multiple trials of simulation for baseline and proposed strategies to validate the efficiency under different conditions. Each simulation runs for at most 1000 iterations due to time constraints.  

To compare the performance of the proposed approach with the baseline strategy, we also need to remove the influence of deadlock. Therefore, in the experiment, we disabled the collision avoidance function of Robotarium (Proposed w/o deadlock avoidance).  The result can be seen in the table \ref{table:results} that shows the proposed approach's effectiveness in terms of mission performance (treasure collected) and energy performance (keeping the robots alive).

Table \ref{table:results} delineates the visible supremacy of the proposed approach in terms of keeping robots alive and values of treasure collected. Embedding deadlock avoidance with the proposed approach helped keep more robots alive and achieve a high value of treasure. However, the proposed approach has traveled more than baseline, spent more time recharging, but waited less on recharging, which is ultimately justified by the achieved treasures picked up that are 15 counts more than the baseline and five counts more than without deadlock strategy.

The proposed approach also attempts to keep the robots connected by applying a maximum connectivity graph algorithm. To elaborate on the significance of graph connection, we have performed some experiments by excluding the graph connection aspect of the approach. Table \ref{table:result1} shows that keeping the graph connected helps the robots to keep alive, resulting in a minor improvement in the task performance.

\section{Conclusion}
\label{sec:conclusion}
This paper provides a very effective energy-aware distributed task allocation algorithm for continuous tasks to be used for cooperative robots. It provided a theoretical and algorithmic explanation of the proposed approach along with the experimental design strategy. In a variety of circumstances, we assessed the proposed solution compared against a typical greedy-based baseline strategy (assigning the closest collection bin to the available robot and recharging the robot to maximum capacity). In comparison with the greedy approach, the proposed approach displayed a considerable improvement in performance and efficiency. 

In the near future, we intend to validate the approach on real swarm robots and extend the idea to other domains of distributed multi-robot systems.
\bibliography{references}
\bibliographystyle{IEEEtran}

\end{document}